# Speeding up the binary Gaussian process classification

**Jarno Vanhatalo**
Department of Biomedical Engineering
and Computational Science
Aalto University
Finland

**Aki Vehtari**
Department of Biomedical Engineering
and Computational Science
Aalto University
Finland

## Abstract

Gaussian processes (GP) are attractive building blocks for many probabilistic models. Their drawbacks, however, are the rapidly increasing inference time and memory requirement alongside increasing data. The problem can be alleviated with compactly supported (CS) covariance functions, which produce sparse covariance matrices that are fast in computations and cheap to store. CS functions have previously been used in GP regression but here the focus is in a classification problem. This brings new challenges since the posterior inference has to be done approximately. We utilize the expectation propagation algorithm and show how its standard implementation has to be modified to obtain computational benefits from the sparse covariance matrices. We study four CS covariance functions and show that they may lead to substantial speed up in the inference time compared to globally supported functions.

## 1 Introduction

Gaussian processes (GP) are powerful tools for many probabilistic modeling problems. They are flexible models that can be used to set priors over functions (Rasmussen and Williams, 2006). The inconvenience with GP models is the unfavorable $O(n^3)$ scaling in the inference time and $O(n^2)$ in the memory requirements, where $n$ is the size of the training set.

The key components in GP are the mean and covariance function, which define, for example, the smoothness of the functions. By using a compactly supported (CS) covariance function the covariance matrix may be sparse, which speeds up the inference and reduces the memory requirements. The idea is not new and CS functions have been utilized especially in spatial statistics (Gneiting, 2002; Wendland, 2005). There the aim is on maximum likelihood estimation and prediction, which collapse into solving sparse linear systems. In the GP literature, the topic has obtained less attention perhaps since there is only little literature on the practical implementation issues, which cover also other techniques than sparse linear solvers. Previously Storkey (1999) and Vanhatalo and Vehtari (2008) have tackled the problem in GP regression.

Here, we consider a classification problem, which is a common example of a GP model with a non-Gaussian likelihood. We utilize the expectation propagation (EP) (Minka, 2001) algorithm for approximate inference and show the steps in its implementation so that the full advantage of the CS functions can be exploited. We compare the performance of four CS covariance functions to the commonly used squared exponential. The results show that we can achieve speed up in the inference time and reduction in the memory requirements with CS functions. The classification model serves as an example of a GP model for which EP is a powerful analysis tool and the techniques described here can easily be applied to other GP models as well.

## 2 Gaussian process binary classification

We will consider a classification problem with binary observations, $y_i \in \{-1, 1\}, i = 1, ..., n$, appointed to inputs $\mathbf{X} = \{\mathbf{x}\}_{i=1}^n$. The observations are considered to be drawn from a Bernoulli distribution with a success probability $p(y_i = 1|\mathbf{x}_i)$. The probability is related to a latent function $f(\mathbf{x}) : \Re^d \to \Re$ that is mapped to a unit interval by a sigmoid transformation. The transformation used is the probit $p(y_i = 1|\mathbf{x}_i) = \Phi(f(\mathbf{x}_i))$, where $\Phi$ denotes the cumulative probability function of the standard Normal density.

The latent function is given a GP prior, which implies that any finite subset of latent variables has a multivariate Gaussian distribution (Rasmussen and

Williams, 2006). Namely, at the observed inputs the latent variables, $\mathbf{f} = \{f(\mathbf{x}_i)\}_{i=1}^n$, have a Gaussian prior distribution $p(\mathbf{f}|\mathbf{X}) = N(\mathbf{f}|\mu, \mathbf{K}_{f,f})$, where $\mathbf{K}_{f,f}$ is the covariance matrix and $\mu$ the mean function. Since neither of the class labels is considered more probable, we set the prior mean to zero. The covariance matrix is constructed by a covariance function $k(\mathbf{x}_i, \mathbf{x}_j|\theta)$, which represents the prior assumptions of the smoothness of the latent function. A widely used covariance function is the stationary squared exponential

$$k_{\text{se}}\left(r|\theta = \{\sigma_{\text{se}}^2, l_1, ..., l_d\}\right) = \sigma_{\text{se}}^2 \exp\left(-r^2\right), \quad (1)$$

where $r = \sqrt{\sum_{d=1}^D (x_{i,d} - x_{j,d})^2 / l_d^2}$ is the distance between two input vectors, and $\sigma_{\text{se}}^2$ the magnitude parameter. The length-scale, $l_d$, governs how fast the correlation decreases among input dimension $d$. The process associated with squared exponential is indefinitely mean square differentiable, which is a very strong assumption on the smoothness of $f$. Nevertheless, the covariance function (1) is probably the most widely used in the machine learning literature.

Given the latent function, the class labels are independent Bernoulli variables and, we can write the conditional posterior of the latent function as

$$p(\mathbf{f}|\mathcal{D}, \theta) = \frac{1}{p(\mathcal{D}|\theta)} p(\mathbf{f}|\mathbf{X}, \theta) \prod_{i=1}^{N} p(y_i|f_i). \quad (2)$$

Here $p(\mathcal{D}|\theta) = \int p(\mathbf{y}|\mathbf{f}) p(\mathbf{f}|\mathbf{X}, \theta) d\mathbf{f}$ is the marginal likelihood of the hyperparameters, and $\mathcal{D} = \{\mathbf{y}, \mathbf{X}\}$. Unfortunately the posterior and the marginal likelihood are analytically intractable.

## 3 Expectation propagation algorithm

Nickisch and Rasmussen (2008) provide an extensive comparison of different methods for approximate inference in GP classification and show that the number one choice in name of accuracy and speed is the expectation propagation (EP) algorithm (Minka, 2001). EP approximates the conditional posterior (2) with

$$q(\mathbf{f}|\mathcal{D}, \theta) = \frac{1}{Z_{\text{EP}}} p(\mathbf{f}|\mathbf{X}, \theta) \prod_{i=1}^{N} t_i(f_i|\tilde{Z}_i, \tilde{\mu}_i, \tilde{\sigma}_i^2), \quad (3)$$

where the likelihood terms have been replaced by *site functions* $t_i(f_i|\tilde{Z}_i, \tilde{\mu}_i, \tilde{\sigma}_i^2) = \tilde{Z}_i N(f_i|\tilde{\mu}_i, \tilde{\sigma}_i^2)$, which are un-normalized Gaussian densities, and the normalizing constant by $Z_{\text{EP}}$. The algorithm tries to match the first two marginal moments of $q(\mathbf{f}|\mathcal{D}, \theta)$, with those of the true posterior. The site terms $\tilde{Z}_i$ are scaling parameters which ensure that also the zeroth moment of the approximate and true posterior match, that is $Z_{\text{EP}} \approx p(\mathcal{D}|\theta, \gamma)$.

The algorithm proceeds as follows. First, we initialize the site parameters $\tilde{Z}_i$, $\tilde{\mu}_i$ and $\tilde{\sigma}_i^2$ and update them sequentially. At each iteration, we first evaluate a cavity distribution $q_{-i}(f_i) = q(f_i|\mathcal{D}, \theta)/t_i(f_i)$, which removes the $i$'th site from the $i$'th marginal posterior. Second step is to find a Gaussian distribution $\hat{q}(f_i)$, which satisfies $\hat{q}(f_i) = \arg\min_q KL\left(q_{-i}(f_i) p(y_i|f_i) || q(f_i)\right)$. This is equivalent to matching the first and second moment between the two distributions (Seeger, 2005). After this, the parameters of the local approximation $t_i$ are updated so that the new marginal posterior $q_{-i}(f_i) t_i(f_i)$ matches with the moments of $\hat{q}(f_i)$. For last the parameters of the approximate posterior (3) are updated: $\boldsymbol{\Sigma} = (\mathbf{K}_{f,f}^{-1} + \tilde{\Sigma}^{-1})^{-1}$ and $\mu = \boldsymbol{\Sigma} \tilde{\Sigma}^{-1} \tilde{\mu}$, where $\tilde{\Sigma} = \text{diag}[\tilde{\sigma}_1^2, ..., \tilde{\sigma}_n^2]$.

Traditionally the posterior covariance is updated with the rank one update (Rasmussen and Williams, 2006)

$$\boldsymbol{\Sigma}^{\text{new}} = \boldsymbol{\Sigma}^{\text{old}} - \mathbf{s}_i \mathbf{s}_i^T \delta_i, \quad (4)$$

$$\text{where } \delta_i = \frac{\tilde{\tau}_i^{\text{new}} - \tilde{\tau}_i^{\text{old}}}{1 + (\tilde{\tau}_i^{\text{new}} - \tilde{\tau}_i^{\text{old}}) \boldsymbol{\Sigma}_{ii}^{\text{old}}},$$

$\tilde{\tau}_i = \tilde{\sigma}_i^{-2}$, and $\mathbf{s}_i$ is the $i$'th column of $\boldsymbol{\Sigma}^{\text{old}}$. This update requires $O(n^2)$ time and since it has to be done for each site, the total computational cost is $O(n^3)$.

### 3.1 Inferring the hyperparameters

The normalization constant $Z_{\text{EP}}$ is EP's approximation for the marginal likelihood, and its logarithm is (Rasmussen and Williams, 2006)

$$\log Z_{\text{EP}} = \frac{-1}{2}\log|\mathbf{K}_{f,f} + \tilde{\Sigma}| - \frac{1}{2}\tilde{\mu}^T\left(\mathbf{K}_{f,f} + \tilde{\Sigma}\right)^{-1}\tilde{\mu} + C \quad (5)$$

where $C$ collects all the terms that are not explicitly dependent on $\theta$. This can be differentiated with respect to the hyperparameters at the fixed point solution as (Seeger, 2005)

$$\frac{\partial \log Z_{\text{EP}}}{\partial \theta} = \frac{1}{2}\tilde{\mu}^T\left(\mathbf{K}_{f,f} + \tilde{\Sigma}\right)^{-1}\frac{\partial \mathbf{K}_{f,f}}{\partial \theta}\left(\mathbf{K}_{f,f} + \tilde{\Sigma}\right)^{-1}\tilde{\mu}$$
$$- \frac{1}{2}\text{tr}\left((\mathbf{K}_{f,f} + \tilde{\Sigma})^{-1}\frac{\partial \mathbf{K}_{f,f}}{\partial \theta}\right). \quad (6)$$

By giving a prior for the hyperparameters, $p(\theta)$, we can search their posterior mode by maximizing $\log Z_{\text{EP}} + \log p(\theta)$. To ensure numerical stability the equations (5) and (6) are transformed so that they utilize the Cholesky decomposition of $\mathbf{B} = \mathbf{I} + \tilde{\Sigma}^{-1/2} \mathbf{K}_{f,f} \tilde{\Sigma}^{-1/2}$.

## 4 Compactly supported covariance functions

With a compactly supported covariance function we mean a function that gives zero correlation between data points whose distance exceeds a certain threshold. The challenge in constructing CS covariance functions is to guarantee their positive definiteness since a covariance function with global support can not be cut arbitrarily while keeping it positive definite. One option is to use a family of piecewise polynomial functions $k_{pp,q}$ such as (Wendland, 2005):

$$k_{pp,0}(r) = \sigma^2 (1-r)_+^j \qquad (7)$$

$$k_{pp,1}(r) = \sigma^2 (1-r)_+^{j+1}((j+1)r+1) \qquad (8)$$

$$k_{pp,2}(r) = \frac{\sigma^2}{3}(1-r)_+^{j+2}((j^2+4j+3)r^2+(3j+6)r+3) \qquad (9)$$

$$k_{pp,3}(r) = \frac{\sigma^2}{15}(1-r)_+^{j+3}((j^3+9j^2+23j+15)r^3+ (6j^2+36j+45)r^2+(15j+45)r+15) \qquad (10)$$

where $j = \lfloor D/2 \rfloor + q + 1$. These functions correspond to processes that are $q$ times mean square differentiable and are positive definite up to dimension $D$. Thus, the smallest possible degree of the polynomial increases as a function of an input dimension. Also the rate of decrease of correlation between two inputs increases as a function of the input dimension since the cutting function $(1-r)_+$ is raised to the power of $j$. Figure 1 illustrates this for input dimensions 2, 5 and 10. The figure, shows also the $k_{se}$ function whose rate of decrease does not depend on input dimension. The CS functions are rougher than $k_{se}$. The $k_{pp,3}(r)$ and $k_{pp,2}(r)$ functions should, however, be smooth enough for most real world applications since they correspond to the mean square differentiability of commonly used Mátern covariance functions.

The computational speed-up and memory savings are the greater the sparser the covariance matrix is. From Figure 1 we can see that as the input dimension increases the length-scale of a piece-wise polynomial covariance function has to increase in order for the function to capture correlations from the same distance as with smaller $D$. However, the cut-off distance is always $r = 1$ and if the length-scale increases the cut-off distance increases as well and the covariance matrix becomes denser. This is illustrated in Figure 2. For each $k_{pp,q}$, we simulated ten data sets from GP with $k_{pp,q}(\mathbf{x}_i, \mathbf{x}_j) + 0.04\mathbf{I}$ covariance function with $D = 2$, where inputs were drawn randomly from $[0\ 10]^2$, and trained a separate GP with the same covariance function but a different $D$ in steps of 5 up to $D = 70$. Figure 2 shows the resulting posterior mode for the length-scale and the density of the covariance matrix with 95% quantiles across the data sets. It can be seen that both of them increase as $D$ increases. The model performance remained the same for each $D$.

The above example illustrates that the piece-wise polynomial covariance functions have a natural characteristics to give denser covariance matrices as $D$ increases, for which reason we should always use as small $D$ as possible. The problem seems to be less severe for smoother functions, with higher $q$. The other disadvantage of increasing input dimension is that the data is more sparsely distributed and it becomes increasingly hard to infer short length-scale phenomena. Thus, one would assume the piece-wise polynomial functions to work best for low dimensional data sets.

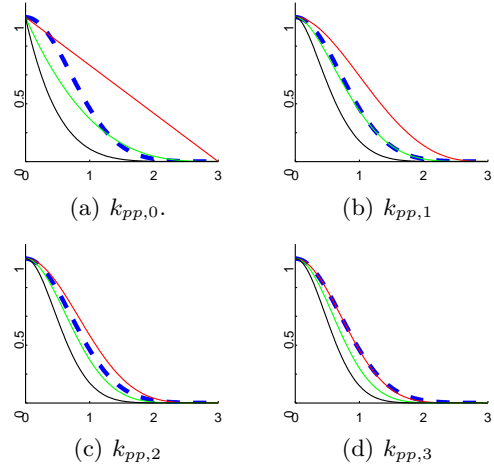

Figure 1: Covariance functions $k_{se}$ (dashed line) and $k_{pp,q}$ with $D = 1$ (red) $D = 5$ (green) and $D = 10$ (black). The length-scales are $l_{se} = 1$ and $l_{pp} = 3$.

Here, we restrict our analysis to the four CS covariance functions above but are aware that there are vast amount of other options also. For example, we could truncate any globally supported covariance function by multiplying it with one of the above functions.

### 4.1 Computations with CS functions

The key role is played by a sparse Cholesky factorization. The Cholesky factorization of a full matrix requires time $O(n^3)$, but for sparse matrices this is faster since the sparsity is retained in the Cholesky factorization. The factorization time depends on the number of symbolically non-zero elements in the Cholesky factorization. Those are elements which have to be modified during the Cholesky factorization even if they were zero at the final result (Davis, 2006). The number of non-zeros in the Cholesky factorization can be reduced by permuting the columns and rows of matrix to be factorized (e.g. Amestoy et al., 2004; Davis, 2006).

The matrix with central importance in the EP algorithm is $\mathbf{B} = \mathbf{I} + \tilde{\Sigma}^{-1/2} \mathbf{K}_{f,f} \tilde{\Sigma}^{-1/2}$ whose sparsity structure is the same as that of $\mathbf{K}_{f,f}$. After finding the sparse Cholesky factorization $\mathbf{L}\mathbf{L}^T = \mathbf{B}$ we can effi-ciently evaluate the log likelihood (5) and its derivatives (6). For example $\log |\mathbf{B}| = 2 \sum_{i=1}^{n} \log \mathbf{L}_{ii}$ and $\mathbf{a}^{\mathrm{T}} \mathbf{B}^{-1} \mathbf{a}$, where $\mathbf{a}$ is an arbitrary vector, is evaluated by first solving the sparse linear equation $\mathbf{L}\mathbf{L}^{\mathrm{T}} \mathbf{v} = \mathbf{a}$ and then evaluating the sparse dot product $\mathbf{a}^{\mathrm{T}} \mathbf{v}$.

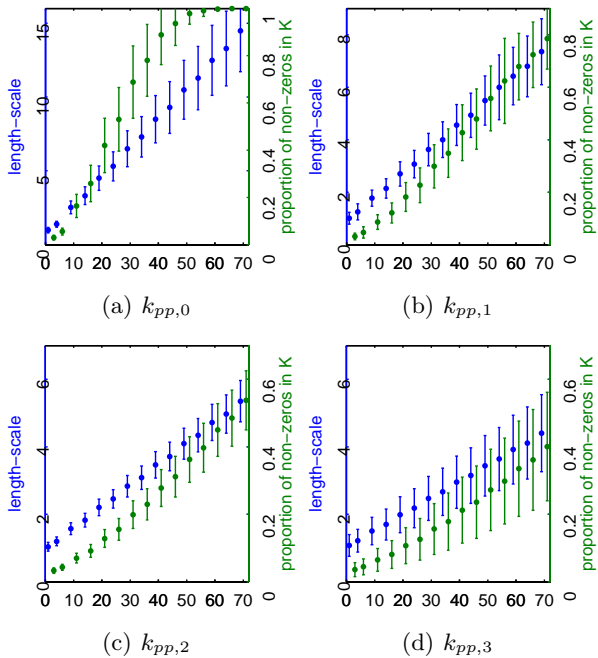

Figure 2: The posterior of the length-scale and the sparsity of the covariance matrix as a function of $d$ for piece-wise polynomial functions. See text for details.

The term that needs most concern is $\mathrm{tr}(\tilde{\Sigma}^{-1/2} \mathbf{B}^{-1} \tilde{\Sigma}^{-1/2} \partial \mathbf{K}_{\mathrm{f,f}} / \partial \theta)$, which occurs in the gradients of the approximate log marginal likelihood (6). If evaluated by first inverting $\mathbf{B}$, this would scale as $O(n^3)$ in time, since the inverse of a sparse matrix is not, in general, sparse. However, we can save substantially time by evaluating only a sparsified inverse. The element $ij$ of a matrix $\partial \mathbf{K}_{\mathrm{f,f}} / \partial \theta$ can be non-zero only if the corresponding element $\mathbf{B}_{ij} \neq 0$. Thus, if we denote $\mathbf{Z} = \tilde{\Sigma}^{-1/2} \mathbf{B}^{-1} \tilde{\Sigma}^{-1/2}$ we can write

$$\mathrm{tr}\left( \mathbf{Z} \frac{\partial \mathbf{K}_{\mathrm{f,f}}}{\partial \theta} \right) = \sum_{i=1}^{n} \sum_{j \in V_i} \mathbf{Z}_{ij}^{\mathrm{sp}} \left[ \frac{\partial \mathbf{K}_{\mathrm{f,f}}}{\partial \theta} \right]_{ij}, \quad (11)$$

where $V_i = \{j | \mathbf{B}_{ij} \neq 0\}$, and $\mathbf{Z}^{\mathrm{sp}}$ is the sparsified representation of $\mathbf{Z}$, which has non-zero elements only, where $\mathbf{B}_{ij} \neq 0$ (Vanhatalo and Vehtari, 2008). $\mathbf{Z}^{\mathrm{sp}}$ can be obtained by using an algorithm introduced by Takahashi et al. (1973) in only a fraction of time required to find the full inverse.

With the above considerations the evaluation of the log marginal likelihood and its derivatives is considerably faster than with full covariance matrix whenever we have successfully run the EP iterations to convergence. However, the most problematic part is the implementation of the iterative algorithm itself.

## 5 Speeding up EP iterations

### 5.1 Updates of posterior moments

The central part in speeding up the EP algorithm with CS covariance functions is to replace the rank one update of the covariance matrix (4) with a Cholesky update of $\mathbf{B}$ and writing the posterior covariance as

$$\boldsymbol{\Sigma} = \mathbf{K}_{\mathrm{f,f}} - \mathbf{K}_{\mathrm{f,f}} \tilde{\Sigma}^{-1/2} \mathbf{L}^{-\mathrm{T}} \mathbf{L}^{-1} \tilde{\Sigma}^{-1/2} \mathbf{K}_{\mathrm{f,f}}, \quad (12)$$

The Cholesky update is in general less prone to numerical problems than the rank one update of the covariance matrix (Seeger, 2008) and the matrix $\mathbf{B}$ is better conditioned than $\boldsymbol{\Sigma}$ (Rasmussen and Williams, 2006). $\mathbf{B}$ will remain sparse throughout the algorithm, whereas the posterior covariance $\boldsymbol{\Sigma} = (\mathbf{K}_{\mathrm{f,f}}^{-1} + \tilde{\Sigma}^{-1})^{-1}$ will not. Thus, if we utilized the rank one update (4) we would be processing full matrix at the end of the EP algorithm.

At each iteration we need only the marginal posterior mean $\mu_i$ and variance $\sigma_i^2$, which can be evaluated efficiently as follows. First we evaluate a sparse vector $\mathbf{a} = \tilde{\Sigma}^{-1/2} [\mathbf{K}_{\mathrm{f,f}}]_{:,i}$ and solve a linear equation $\mathbf{L}\mathbf{L}^{\mathrm{T}} \mathbf{a} = \mathbf{t}$. After this we can evaluate the marginal variance as $\sigma_i^2 = [\mathbf{K}_{\mathrm{f,f}}]_{i,i} - \mathbf{a}^{\mathrm{T}} \mathbf{t}$. The posterior mean can be written as $\mu = \gamma - \mathbf{K}_{\mathrm{f,f}} \tilde{\Sigma}^{-1/2} \mathbf{L}^{-\mathrm{T}} \mathbf{L}^{-1} \tilde{\Sigma}^{-1/2} \gamma$, where $\gamma = \mathbf{K}_{\mathrm{f,f}} \tilde{\Sigma}^{-1} \tilde{\mu}$. If we denote $\tilde{\nu} = \tilde{\Sigma}^{-1} \tilde{\mu}$ we can update $\gamma$ at each iteration as $\gamma^{\mathrm{new}} = \gamma^{\mathrm{old}} + [\mathbf{K}_{\mathrm{f,f}}]_{:,i} \Delta \tilde{\nu}_i$, where $\Delta \tilde{\nu}_i = \tilde{\nu}_i^{\mathrm{new}} - \tilde{\nu}_i^{\mathrm{old}}$. The marginal mean can then be evaluated as $\mu_i = \gamma_i - \mathbf{t}^{\mathrm{T}} (\tilde{\Sigma}^{-1/2} \gamma)$.

Evaluating $\mathbf{a}$ and updating $\gamma$ scales as $O(\mathrm{nnz}([\mathbf{K}_{\mathrm{f,f}}]_{:,i}))$, where $\mathrm{nnz}(\cdot)$ stands for the number of nonzero elements. Since $\mathbf{a}$ is sparse, we can solve $\mathbf{t}$ efficiently in time proportional to the time needed to multiply $\mathbf{L}$ times $\mathbf{t}$, which is $O(\sum_{\mathbf{t}_j \neq 0} \mathrm{nnz}((\mathbf{L})_{:j}))$ (Davis, 2006). The sparse dot product $\mathbf{a}^{\mathrm{T}} \mathbf{t}$ scales as $O(\max[\mathrm{nnz}(\mathbf{a}), \mathrm{nnz}(\mathbf{t})])$. The pseudo-code for the sparse EP algorithm is shown in algorithm 1, where the only remaining question is the update procedure for $\mathbf{L}$, *ldlrowmodify*, which will be considered next.

### 5.2 The row modification algorithm

At each iteration, we first update the $i$'th diagonal of $\tilde{\Sigma}$ which in turn leads to an update in the $i$'th column and the $i$'th row of the matrix $\mathbf{B}$. The computational savings can be achieved since these changes affect only the $i$'th row and the columns $k \geq i$ of the Cholesky decomposition of $\mathbf{B}$ (Davis and Hager, 2005). By updating only the terms that change, the Cholesky up-dates become increasingly fast as the EP algorithm proceeds. This is illustrated with an LDL Cholesky factorization

$$\mathbf{B} = \begin{bmatrix} \mathbf{B}_{11} & \mathbf{b}_{12} & \mathbf{B}_{31}^{\text{T}} \\ \mathbf{b}_{12}^{\text{T}} & b_{22} & \mathbf{b}_{32}^{\text{T}} \\ \mathbf{B}_{31} & \mathbf{b}_{32} & \mathbf{B}_{33} \end{bmatrix}$$
$$= \begin{bmatrix} \mathbf{L}_{11} & & \\ \mathbf{l}_{12}^{\text{T}} & 1 & \\ \mathbf{L}_{31} & \mathbf{l}_{32} & \mathbf{L}_{33} \end{bmatrix} \begin{bmatrix} \mathbf{D}_{11} & & \\ & d_{22} & \\ & & \mathbf{D}_{33} \end{bmatrix} \begin{bmatrix} \mathbf{L}_{11}^{\text{T}} & \mathbf{l}_{12} & \mathbf{L}_{31}^{\text{T}} \\ & 1 & \mathbf{l}_{32}^{\text{T}} \\ & & \mathbf{L}_{33}^{\text{T}} \end{bmatrix}, \quad (13)$$

where the vector $[\mathbf{b}_{12}^{\text{T}} \ b_{22} \ \mathbf{b}_{32}^{\text{T}}]$ corresponds to the $i$'th row of $\mathbf{B}$. The LDL Cholesky decomposition can be transformed to regular Cholesky decomposition by multiplying lower triangular $\mathbf{L}$ by $\mathbf{D}^{1/2}$ from right. The change in $\mathbf{b}_{12}$, $b_{22}$ and $\mathbf{b}_{32}$ lead to change in terms $\mathbf{L}_{33}$, $\mathbf{l}_{12}^{\text{T}}$, $\mathbf{l}_{32}$, $\mathbf{D}_{33}$ and $d_{22}$ of the Cholesky decomposition. The new Cholesky decomposition is

$$\bar{\mathbf{B}} = \begin{bmatrix} \mathbf{B}_{11} & \bar{\mathbf{b}}_{12} & \mathbf{B}_{31}^{\text{T}} \\ \bar{\mathbf{b}}_{12}^{\text{T}} & \bar{b}_{22} & \bar{\mathbf{b}}_{32}^{\text{T}} \\ \mathbf{B}_{31} & \bar{\mathbf{b}}_{32} & \mathbf{B}_{33} \end{bmatrix}$$
$$= \begin{bmatrix} \mathbf{L}_{11} & & \\ \bar{\mathbf{l}}_{12}^{\text{T}} & 1 & \\ \mathbf{L}_{31} & \bar{\mathbf{l}}_{32} & \bar{\mathbf{L}}_{33} \end{bmatrix} \begin{bmatrix} \mathbf{D}_{11} & & \\ & \bar{d}_{22} & \\ & & \bar{\mathbf{D}}_{33} \end{bmatrix} \begin{bmatrix} \mathbf{L}_{11}^{\text{T}} & \bar{\mathbf{l}}_{12} & \mathbf{L}_{31}^{\text{T}} \\ & 1 & \bar{\mathbf{l}}_{32}^{\text{T}} \\ & & \bar{\mathbf{L}}_{33}^{\text{T}} \end{bmatrix}, \quad (14)$$

where the changed terms are denoted by overlying bar.

**Algorithm 1** Pseudo-code for the EP algorithm. ldl-rowmodify is a Cholesky row-modification algorithm discussed in section 5.3
**Input:** $\mathbf{K}_{f,f}$, $\mathbf{y}$
 1: initialize $\tilde{\nu}$ and $\tilde{\tau}$
 2: evaluate $\mathbf{L} = \text{chol}(\mathbf{B})$
 3: **while** $|\Delta \log Z_{\text{EP}}| > tol$ **do**
 4:    **for** $i = 1$ to $n$ **do**
 5:       $\mathbf{a} = \tilde{\Sigma}^{-1/2}[\mathbf{K}_{f,f}]_{:,i}$
 6:       $\mathbf{t} = (\mathbf{L}^{\text{T}} \backslash (\mathbf{L} \backslash \mathbf{a}))$
 7:       $\sigma_i^2 = [\mathbf{K}_{f,f}]_{i,i} - \mathbf{a}^{\text{T}} \mathbf{t}$
 8:       $\mu_i = \gamma_i - \mathbf{t}^{\text{T}}(\tilde{\Sigma}^{-1/2} \gamma)$
 9:       evaluate $\nu_i^{\text{new}}$ and $\tilde{\tau}^{\text{new}}$ as in (Rasmussen and Williams, 2006)
10:       set $\tilde{\Sigma}_{i,i} = \tilde{\tau}_i^{-1}$ and $\mathbf{a}_i = \sqrt{\tilde{\tau}_i}[\mathbf{K}_{f,f}]_{i,i}$
11:       $\mathbf{d} = \mathbf{a}\tilde{\sigma}_i + \mathbf{e}_i$       (this is $[\mathbf{B}^{\text{new}}]_{:,i}$)
12:       $\mathbf{L} = \text{ldlrowmodify}(\mathbf{L}, \mathbf{d})$
13:       $\gamma = \gamma + [\mathbf{K}_{f,f}]_{:,i}\Delta\tilde{\nu}_i$
14:    **end for**
15:    compute $\log Z_{\text{EP}}^{\text{new}}$ and $\Delta \log Z_{\text{EP}}$
16: **end while**
17: **return** $\mathbf{L}, \mu, \log Z_{\text{EP}}^{\text{new}}$

The Cholesky decomposition can be updated with successive usage of row deletion and row addition algorithms by Davis and Hager (2005). The row deletion updates $\mathbf{L}$ to correspond the factorization of $\mathbf{B}$ where $\bar{\mathbf{b}}_{12}$ and $\bar{\mathbf{b}}_{32}$ are set to zero and $\bar{b}_{22}$ to one. The row addition algorithm, on its part, updates $\mathbf{L}$ to correspond $\mathbf{B}$ where $\bar{\mathbf{b}}_{12}$ and $\bar{\mathbf{b}}_{32}$ are updated from zero to non-zero and $\bar{b}_{22}$ to other than one. Thus, the row deletion corresponds to setting $\tilde{\Sigma}_{ii}^{-1} = 0$, evaluating $\mathbf{B}$ and re-evaluating its Cholesky factorization. Similarly the row addition algorithm corresponds to setting $\tilde{\Sigma}_{ii}^{-1} = \tilde{\tau}_i^{\text{new}}$, evaluating $\mathbf{B}$ a second time, and re-evaluating $\mathbf{L}$.

If we consider a full matrix $\mathbf{B}$, the row deletion algorithm performs $O(2(n-k))$ and the row addition algorithm $O(2n^2 + k^2 - 2nk)$ operations, where $k$ is the row to be modified. Thus, the total number of operations during one sweep of EP algorithm with full covariance matrix still scales as $O(n^3)$. If the covariance matrix is sparse, the computational savings can be substantial, since only the non-zero elements are updated. The general algorithm of Davis and Hager (2005) assumes that the sparsity structure of $\mathbf{B}$ might change during the row addition/deletion. However, if $\tilde{\tau}$ and $\tilde{\nu}$ are non-zero before and after the update the sparsity structure of $\mathbf{B}$ does not change, for which reason we can save more time by performing the deletion and addition steps simultaneously. With the general algorithm we would need to re-analyze the sparsity structure at each iteration and perform unnecessary evaluations that would first decrease the number of non-zeros in the Cholesky factorization and then increase it back to its original state. The special version of the algorithm is discussed next.

### 5.3 Row modifications of B

Following Davis and Hager (2005), we can solve the new elements of the Cholesky decomposition as follows. From (14) we get

$$\mathbf{L}_{11}\mathbf{D}_{11}\bar{\mathbf{l}}_{12} = \bar{\mathbf{b}}_{12}.$$

Since $\bar{\mathbf{b}}_{12}$ is sparse we can solve $\bar{\mathbf{l}}_{12}$ efficiently in time proportional to the time needed to multiply $\mathbf{L}_{11}$ times $\bar{\mathbf{l}}_{12}$, which is $\sum_{(\bar{\mathbf{l}}_{12})_j \neq 0} \text{nnz}((\mathbf{L}_{11})_{:j})$ (Davis, 2006). Computing $\bar{d}_{22}$ can be computed in time proportional to $\text{nnz}(\mathbf{l}_{12})$ from the relation

$$\bar{\mathbf{l}}_{12}^{\text{T}}\mathbf{D}_{11}\bar{\mathbf{l}}_{12} + \bar{d}_{22} = \bar{b}_{22}.$$

Similarly the $i$'th column $\bar{\mathbf{l}}_{32}$ can be solved from

$$\mathbf{L}_{31}\mathbf{D}_{11}\bar{\mathbf{l}}_{12} + \bar{\mathbf{l}}_{32}\bar{d}_{22} = \bar{\mathbf{b}}_{32}.$$

Here, the key component is the matrix vector multiplication $\mathbf{L}_{31}\mathbf{D}_{11}\bar{\mathbf{l}}_{12}$, which scales as $O(\sum_{(\bar{\mathbf{l}}_{12})_j \neq 0} \text{nnz}((\mathbf{L}_{31})_{:j}))$.

The remaining terms can be solved with rank one update and downdate of the Cholesky decomposition (Davis and Hager, 2005)

$$\bar{\mathbf{L}}_{33}\bar{\mathbf{D}}_{33}\bar{\mathbf{L}}_{33}^{\text{T}} = \mathbf{L}_{33}\mathbf{D}_{33}\mathbf{L}_{33}^{\text{T}} + \mathbf{w}_1\mathbf{w}_1^{\text{T}} - \mathbf{w}_2\mathbf{w}_2^{\text{T}}, \quad (15)$$

where $\mathbf{w}_1 = \mathbf{l}_{32}\sqrt{d_{22}}$ and $\mathbf{w}_2 = \bar{\mathbf{l}}_{32}\sqrt{\bar{d}_{22}}$. Now, since the sparsity structure of the Cholesky factorization does not change we can perform the update and downdate simultaneously. This saves some time compared to performing first downdate and then update since the data structure for $\bar{\mathbf{L}}_{33}$ need not be scanned. The number of floating point operations will, however, be the same in either way.

The computational time is dominated by the Cholesky update of $\bar{\mathbf{L}}_{33}$ at the early stage of the algorithm. As the algorithm proceeds the total amount of work will be significantly reduced since the sparse matrix vector multiplications are much lighter operations than the rank one Cholesky update and downdate. The pseudo-code for the row updates is given in algorithm 2.

---

**Algorithm 2** Pseudo-code for ldlrowmodify, the Cholesky update after row modifications to $\mathbf{B}$.

**Input:** $\mathbf{L}$, $[\bar{\mathbf{b}}_{12}^\mathrm{T}\ \bar{b}_{22}\ \bar{\mathbf{b}}_{32}^\mathrm{T}]^\mathrm{T}$, $i$
1: divide $\mathbf{L}$ into LDL form
2: $\bar{\mathbf{l}}_{12} = \mathbf{D}_{11}^{-1}(\mathbf{L}_{11}\backslash\bar{\mathbf{b}}_{12})$
3: $\bar{\mathbf{l}}_{32} = (\bar{\mathbf{b}}_{32} - \mathbf{L}_{31}\mathbf{D}_{11}\bar{\mathbf{l}}_{12})/\bar{d}_{22}$
4: set $\mathbf{w}_1 = \mathbf{l}_{32}\sqrt{d_{22}}$ and $\mathbf{w}_2 = \bar{\mathbf{l}}_{32}\sqrt{\bar{d}_{22}}$
5: conduct Cholesky update and downdate to get $\bar{\mathbf{L}}_{33}\bar{\mathbf{D}}_{33}\bar{\mathbf{L}}_{33} = \mathbf{L}_{33}\mathbf{D}_{33}\mathbf{L}_{33} + \mathbf{w}_1\mathbf{w}_1^\mathrm{T} - \mathbf{w}_2\mathbf{w}_2^\mathrm{T}$
6: set $\mathbf{L}_{i,1:i-1} = \bar{\mathbf{l}}_{12}$, $\mathbf{L}_{i+1:n,i} = \bar{\mathbf{l}}_{32}$, $\mathbf{D}_{ii} = \bar{d}_{22}$, $\mathbf{L}_{i+1:n,i+1:n} = \bar{\mathbf{L}}_{33}$, and $\mathbf{D}_{i+1:n,i+1:n} = \bar{\mathbf{D}}_{33}$,
7: Evaluate $\mathbf{L} = \mathbf{L}\mathbf{D}^{1/2}$
8: **return L**

---

### 5.4 Computational complexity

The computational complexity of the EP algorithm is dominated by the row modifications of the Cholesky decomposition and solving the vector $\mathbf{t}$. It depends on the number of non-zero elements in the columns of $\mathbf{L}$. Since the sparsity of $\mathbf{L}$ is data dependent, exact time scalings cannot be provided. However, we can study a simplified situation.

Let $\eta_i$ denote the number of non-zero elements in the column $i$ of $\mathbf{L}$. Solving $\mathbf{t}$ scales as $O(\sum_{\mathbf{t}_j\neq 0}\eta_j)$, which depends on both $\eta_j$ and the sparsity of $\mathbf{t}$. If $\mathbf{t}$ is full vector and the proportion of nonzero elements per column remains approximately constant, the reduction in time compared to a full matrix will be $\eta_j/j \approx p$, where $p$ is the approximately constant ratio. If $\mathbf{t}$ also is sparse, the time for solving $\mathbf{t}$ will scale down by $\mathrm{nnz}(\mathbf{t})p/n < 1$. Thus already with $p = \mathrm{nnz}(\mathbf{t})/n = 0.5$ the computational time will be one fourth of that required with full covariance matrix.

The most time consuming operations in the Cholesky update after row modifications to $\mathbf{B}$ are solving for $\bar{\mathbf{l}}_{12}$, evaluating $\mathbf{L}_{31}\mathbf{D}_{11}\bar{\mathbf{l}}_{12}$ and performing rank one update and downdate for $\mathbf{L}_{33}$. The first two operations scale as $O(\sum_{(\bar{\mathbf{l}}_{12})_j\neq 0}\mathrm{nnz}((\mathbf{L}_{11})_{:j}) + \sum_{(\bar{\mathbf{l}}_{12})_j\neq 0}\mathrm{nnz}((\mathbf{L}_{31})_{:j})) = O(\sum_{(\bar{\mathbf{l}}_{12})_j\neq 0}\eta_j)$. The same consideration as above applies also here but now the length of $(\bar{\mathbf{l}}_{12})_j$ varies between iterations and is in average $1/n\sum_{i=1}^n i = (n+1)/2$. Thus, these two operations require only about half the time of that to solve $\mathbf{t}$ if the covariance matrix is full. The update of $\mathbf{L}_{33}$ is optimal in the sense that it requires time proportional to the number of non-zero elements in $\mathbf{L}_{33}$, which is in average $1/n\sum_{i=1}^n\sum_{k=i+1}^n \eta_k$. With full covariance matrix this gives time scaling $O(1/n\sum_{i=1}^n i(i+1)/2) = O(n^2)$ and if $\eta_i = p \times i$ this is $O(p^2n^2)$, which scales down the time substantially compared to the full matrix case if $p$ is small. It is seen also that with a full covariance matrix our implementation of EP scales similarly to the traditional one.

We assumed above that the number of non-zeros per column, $\eta$, scales with constant ratio to the number of nonzeros in the Cholesky decomposition of full covariance matrix. This is not the case, in general, but even if the columns had different ratio, $p$, the average of the ratios should reflect the computational complexity. Thus, the ratio of the number of nonzero elements in $\mathbf{L}$ to the number of non-zero elements in the Cholesky decomposition of full covariance matrix, $(n(n+1)/2)$, can be used to approximate the computational savings.

Rasmussen and Williams (2006) suggest to initialize the $\tilde{\nu}$ and $\tilde{\tau}$ to zero. In this case, we cannot use the ldlrowmodify algorithm at the first iteration but we have to use the row addition algorithm. This does not increase the inference time since the Cholesky decomposition of $\mathbf{B}$ is extensively sparse at the early iterations and thus analyzing the new sparsity structure of $\mathbf{L}$ is cheap. In fact the first round of EP is faster with $\tilde{\nu}$ and $\tilde{\tau}$ initialized to zero than to non-zero.

## 6 Experiments

### 6.1 Simulation studies

Here we study the scaling of the EP algorithm as the data size increases. We constructed two data sets by sampling 15 000 inputs randomly from the hypercubes $[0\ 10]^2$ and $[0\ 10]^5$. After this we drew 200/1000 center points which were assigned randomly to either class. Then each input was assigned to the class of its nearest center point. The GP model was then trained with subsets of sizes 500, 1000, 2000, 5000, and 10 000 and tested with the remaining 5000 test points. The number of clusters may seem large but in practice most of the neighboring cluster centers share the same class and the class boundaries vary smoothly. The number of cluster centers is chosen so that the covariance

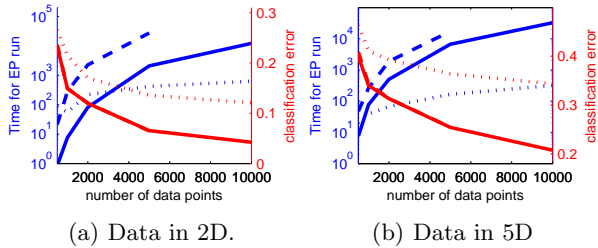

(a) Data in 2D.  (b) Data in 5D

Figure 3: EP's running times and classification errors for $k_{se}$ (dashed line), $k_{pp,3}$ (solid line) and FIC (dotted line) models.

matrix of CS functions is sparse.

As a baseline covariance function we use $k_{se}$ since it is the most commonly used in the GP literature. We compare our methods also to one of the sparse approximations to GP, namely the fully independent conditional (FIC) (Snelson and Ghahramani, 2006; Quiñonero-Candela and Rasmussen, 2005). We chose FIC since it has been shown to perform well in classification tasks and there is a detailed description of its implementation for EP (Naish-Guzman and Holden, 2008). FIC is based on introducing an additional set of latent variables and inputs appointed to them. These inducing inputs are considered hyperparameters that are optimized alongside $\theta$. FIC's computational time scales as $O(nm^2)$, where $m$ is the number of the inducing inputs. Speed up is achieved whenever $m \ll n$. We implemented EP following the lines of Rasmussen and Williams (2006) for $k_{se}$ and Naish-Guzman and Holden (2008) for FIC. We used $m = 400$ inducing inputs, since the underlying latent phenomena are fast varying and less inducing inputs did not work well enough. Optimization was conducted using the scaled conjugate gradient method. The hyperparameters were given a half Student-$t$ prior (Gelman, 2006) with 4 degrees of freedom and scale 6. The prior is weakly informative since it places more mass on small hyperparameter values but due to the heavy tails of Student-$t$ distribution allows posterior to concentrate elsewhere also.

Figure 3 shows the times for a single EP run until its convergence at the posterior mode of the hyperparameters and the classification errors. The CS function used was $k_{pp,3}$. It can be seen that the classification accuracy of $k_{pp,3}$ is identical to the $k_{se}$ covariance function. FIC approximation works less well. With 2 dimensional data, the $k_{pp,3}$ function is about 10–20 and with 5 dimensional data about 3–7 times faster than $k_{se}$. The running time seems to scale as $O(n^3)$ for both the $k_{pp,3}$ and $k_{se}$ function. FIC's running time increases about linearly as it should.

The density of the covariance matrix fill-K = $\text{nnz}(\mathbf{K}_{f,f})/n^2$ and its Cholesky decomposition, fill-L = $\text{nnz}(\mathbf{L})/(n(n+1)/2)$, are summarized in Table 1. fill-L increases in both data sets as the number of data points increases and is higher for the 5 dimensional data as expected. The changes in the number of non-zeros explain also the differences between the speed ups. The training times and the fill-L statistics seem to justify the considerations of the computational complexity in section 5.4. Remarkable point is also that the density of the Cholesky decomposition seems to increase faster than the density of the covariance matrix. The reason for this may be the properties of the AMD ordering algorithm used in the experiments. It is seen also that with 5 dimensional data the density of the covariance matrix first decreases and then starts to increase. Presumably 500 data points in $[0\ 10]^5$ are distributed too sparsely to infer the fast varying latent phenomenon, and the solution is too smooth. This is seen also in the classification errors which are high with small data sets (Figure 3(b)). As $n$ increases the GP model is able to find the fast varying latent phenomenon and the covariance matrix becomes sparser.

Table 1: The density of the covariance matrix and its Cholesky decomposition, fill-L/fill-K, in per cents.

| Data | $n=500$ | $n=1000$ | $n=2000$ | $n=5000$ | $n=10^4$ |
|---|---|---|---|---|---|
| 2D | 12/5 =2.6 | 15/5 =3.2 | 18/5 =3.6 | 19/5 =4.1 | 19/4 =4.3 |
| 5D | 83/36 =2.3 | 72/17 =4.3 | 82/20 =4.0 | 90/23 =3.9 | 96/21 = 4.6 |

FIC is clearly the worst model since the latent phenomena are so fast varying that FIC is not able to capture them. This could be fixed by using even more inducing inputs but then the optimization would be really heavy. The EP's running time does not tell the whole truth for FIC since it does not consider the optimization of parameters. The hyperparameter optimization becomes increasingly hard as we add more inducing inputs and makes the optimization slower. The $k_{pp,2}$ covariance function worked as well as $k_{pp,3}$ but the other CS functions were little inferior to them in their predictive performance.

### 6.2 Experiments with real data

In this section we show results on six classification data sets (see Table 2) chosen from the UCL machine Learning Repository (Asuncion and Newman, 2007). The purpose of the experiments is to demonstrate that the CS functions presented here work also for real data. As in the previous section we compare the CS functions to the $k_{se}$ covariance function and FIC. We used $m = 10$ inducing inputs to keep the optimization as fast as with $k_{pp,3}$. We use rather small data sets so that we are able to analyze them also with the full GP with $k_\text{se}$ function.

Table 2: The description of data set, the classification error (err) and negative log predictive density (nlpd).

| Data set | $n/d$ | $k_\text{se}$ err/nlpd | $k_\text{pp,3}$ err/nlpd | FIC err/nlpd |
|---|---|---|---|---|
| Australian | 690/14 | .13/.32 | .13/.33 | .14/.33 |
| Breast | 683/9 | .03/.99 | .03/.99 | .03/.94 |
| Crabs | 200/6 | .00/.02 | .00/.02 | .00/.02 |
| Ionosphere | 351/33 | .11/.33 | .11/.39 | .20/.41 |
| Pima | 768/8 | .23/.47 | .24/.47 | .24/.48 |
| Sonar | 208/60 | .13/.65 | .01/.69 | .22/.45 |

Table 3: The optimization time (opt), time for a single EP run (EP) and the density of the Cholesky decomposition of the CS covariance matrix, fill-L.

| Data set | fill-L | $k_\text{se}$ opt/EP | $k_\text{pp,3}$ opt/EP | FIC opt/EP |
|---|---|---|---|---|
| Australian | 0.82 | 1170/41 | 900/31 | 250/2 |
| Breast | 1.00 | 1250/80 | 800/45 | 300/2 |
| Crabs | 0.47 | 40/3 | 40/2 | 29/0.3 |
| Ionosphere | 0.90 | 122/7 | 81/4 | 260/1 |
| Pima | 1.00 | 1460/40 | 770/30 | 230/2 |
| Sonar | 0.06 | 26/1 | 2/0.2 | 25/0.2 |

Table 3 summarizes the CPU time needed to optimize all the hyperparameters and the times for single EP run at the posterior mode of the hyperparameters. The table records also the fill-L of the CS covariance function at the posterior mode. The model performance is evaluated with 10-fold cross-validation using the negative log predictive density (nlpd) and the classification error (err), which are summarized in Table 2. The tables summarize the results for $k_{pp,3}$ function. There was no practical difference between $k_{pp,2}$ and $k_{pp,3}$ but $k_{pp,1}$ and $k_{pp,0}$ worked little worse than the other two.

From Table 3 we can see that even if the CS covariance matrix was full at the end of the optimization, the inference time was smaller than with the squared exponential. Thus, we do not lose anything by using CS covariance functions. With sonar and inonosphere data FIC had similar problems that were described with simulated data. The joint optimization of the hyperparameters and inducing inputs converged very slowly. The other models required in average 10-12 optimization steps whereas FIC took always the maximum number of optimization steps, which was 50. Other reason for long optimization times is the much higher number of hyperparameters in FIC. In particular, with 10 inducing inputs, we have $10D$ hyperparameters more, and evaluating gradients will be approximately ten times slower than with only $\theta$ to optimize. However, finding an EP approximation with given hyperparameters is fastest with FIC.

## 7 Discussion

Traditional solutions to computational challenges with GP models have been sparse approximations for a globally supported covariance function. The problems there are how to choose the inducing inputs and that they are not able to model fast varying latent phenomena. CS covariance functions can be utilized just as global covariance functions, without need to tune inducing inputs, and are able to model both local and global correlations. The drawback, for their part, is that the covariance matrix converges to dense if the latent phenomenon is very slow varying. CS covariance functions have probably been less attractive also since they need explicit modifications to matrix routines that are built in functionalities in statistical software packages. However, in this paper we have shown that with very small modifications to these routines we can achieve considerable time savings. The CS covariance functions still scale as $O(n^3)$ in computation time but the constant terms may be considerably smaller than with globally supported functions. Here we demonstrated that the inference scales down approximately at the same rate as the number of zero elements in the Cholesky decomposition of the covariance matrix increase.

The CS covariance functions $k_\text{pp,2}$ and $k_\text{pp,3}$ worked as well as the $k_\text{se}$ in all our experiments. The other two CS functions seem to be little inferior to $k_\text{pp,2}$ and $k_\text{pp,3}$ since they represent rougher processes. An unpleasant property of the piece-wise polynomial functions is that their sparsity properties seem to degrade as the input dimension increases. However, radial CS functions are dependent on $d$ by construction (Wendland, 2005) why one future research direction should be to alleviate the problems from this dependence. Another important topic is a detailed evaluation of different permutation algorithms, since they influence directly the sparsity of the Cholesky decomposition. The implementation issues with CS covariance functions are similar to the computational considerations with Gaussian Markov random fields (GMRF), where the precision matrix is sparse (Rue and Held, 2005). Also with GMRFs the computational benefits are obtained from sparse Cholesky decomposition.

The sparsity of the covariance matrix can be governed by the prior for the length-scale, since the more mass we appoint to small length-scales the more weight we give for sparse covariance matrices. Thus, the Student-$t$ distribution works also as a sparsity prior since it implicitly favors solutions with sparse covariance matrix.

This is not problematic as long as we keep the degrees of freedom small and the scale reasonably large so that the data is able to overrule the prior. It seems also that the length-scale and magnitude are under identifiable and the proportion $\sigma^2/l$ is more important to the predictive performance than their individual values. This property is well known in the spatial statistics community in relation to Mátern covariance functions (e.g. Diggle et al., 1998). Traditionally the priors for covariance function parameters are not given much attention in the ML literature and only the marginal likelihood is maximized (which implies uniform prior for the hyperparameters). However, with CS functions the priors should be given more weight, since it is computationally beneficial to favor sparse covariance matrices.

The CS covariance functions and the techniques discussed here do not replace the sparse approximations. There are still many problems where exact knowledge of the training time is needed and one cannot rely on only possibly sparse covariance matrix. However, there are also many problems where the exact scaling of the inference time is not crucial but one would hope for fast inference. For these problems the CS covariance functions offer a good tool. It should be noticed also that CS covariance functions model local correlations whereas sparse approximations, such as FIC, aim to approximate the global correlations. Thus, these two approaches should not be seen as competing but complementary. One could also begin the data analysis with CS covariance functions and continue with sparse approximations if the length-scale grows too large.

We have considered GP classification but the same techniques can be used in many other GP models with non-Gaussian observations by just modifying the minimization of the KL divergence in EP algorithm. Thus, the benefits of CS covariance functions can be enjoyed with those models as well. The codes used in this work are available in the internet (http://www.lce.hut.fi/research/mm/gpstuff/) as a part of a GP software package, GPstuff.

## Acknowledgements

This research was funded by the Academy of Finland, and the Graduate School in Electronics and Telecommunications and Automation (GETA).

## References

Amestoy, P., Davis, T. A., and Duff, I. S. (2004). Algorithm 837: AMD, an approximate minimum degree ordering algorithm. *ACM Transactions on Mathematical Software*, 30(3):381–388.

Asuncion, A. and Newman, D. (2007). UCI machine learning repository.

Davis, T. A. (2006). *Direct Methods for Sparse Linear Systems*. SIAM.

Davis, T. A. and Hager, W. W. (2005). Row modifications of a sparse Cholesky factorization. *SIAM Journal on Matrix Analysis and Applications*, 26(3):621–639.

Diggle, P. J., Tawn, J. A., and Moyeed, R. A. (1998). Model-based geostatistics. *Journal of the Royal Statistical Society. Series C (Applied Statistics)*, 47(3):299–350.

Gelman, A. (2006). Prior distributions for variance parameters in hierarchical models. *Bayesian Analysis*, 1(3):515–533.

Gneiting, T. (2002). Compactly supported correlation functions. *Journal of Multivariate Analysis*, 83:493–508.

Minka, T. (2001). *A family of algorithms for approximate Bayesian inference*. PhD thesis, Massachusetts Institute of Technology.

Naish-Guzman, A. and Holden, S. (2008). The generalized FITC approximation. In Platt, J., Koller, D., Singer, Y., and Roweis, S., editors, *Advances in Neural Information Processing Systems 20*. MIT Press, Cambridge, MA.

Nickisch, H. and Rasmussen, C. E. (2008). Approximations for binary Gaussian process classification. *Journal of Machine Learning Research*, 9:2035–2078.

Quiñonero-Candela, J. and Rasmussen, C. E. (2005). A unifying view of sparse approximate Gaussian process regression. *Journal of Machine Learning Research*, 6(3):1939–1959.

Rasmussen, C. E. and Williams, C. K. I. (2006). *Gaussian Processes for Machine Learning*. The MIT Press.

Rue, H. and Held, L. (2005). *Gaussian Markov Random Fields Theory and Applications*. Chapman & Hall/CRC.

Seeger, M. (2005). Expectation propagation for exponential families. Technical report, Max Planck Institute for Biological Cybernetics, Tübingen, Germany.

Seeger, M. (2008). Bayesian inference and optimal design for the sparse linear model. *Journal of Machine Learning Research*, 9:759–813.

Snelson, E. and Ghahramani, Z. (2006). Sparse Gaussian process using pseudo-inputs. In Weiss, Y., Schlkopf, B., and Platt, J., editors, *Advances in Neural Information Processing Systems 18*. The MIT Press.

Storkey, A. (1999). *Efficient Covariance Matrix Methods for Bayesian Gaussian Processes and Hopfield Neural Networks*. PhD thesis, University of London.

Takahashi, K., Fagan, J., and Chen, M.-S. (1973). Formation of a sparse bus impedance matrix and its application to short circuit study. In *Power Industry Computer Application Conference Proceedings*. IEEE Power Engineering Society.

Vanhatalo, J. and Vehtari, A. (2008). Modelling local and global phenomena with sparse Gaussian processes. In McAllester, D. A. and Myllymäki, P., editors, *Proceedings of the 24th Conference on Uncertainty in Artificial Intelligence*, pages 571–578.

Wendland, H. (2005). *Scattered Data Approximation*. Cambridge University Press.